\documentclass{sig-alternate}[10pt] 
\usepackage[
  pass,
]{geometry}
\usepackage{mathtools}
\usepackage{mathptmx} 
\usepackage{comment}
\newcommand{\ignore}[1]{}
\usepackage{fancyhdr}
\usepackage[normalem]{ulem}
\usepackage[hyphens]{url}
\usepackage{hyperref}
\usepackage[sort,nocompress]{cite}
\usepackage[final]{microtype}
\usepackage{amsmath}
\usepackage{flushend}
\usepackage[]{subfig}
\usepackage[table]{xcolor}
\usepackage{xspace}
\usepackage{array}
\usepackage{xcolor}
\usepackage{subfig}
\usepackage{float}
\usepackage{amssymb}

\newcommand{\BSD}{\textit{LM}\xspace}

\newcommand{\BASE}{\textit{BASE}\xspace}

\newcommand{\OURLCORE}{Laconic} 
\newcommand{\OURSCORE}{LAC}
\newcommand{\OURL}{\textit{\OURLCORE}\xspace} 
\newcommand{\OURS}{\textit{\OURSCORE}\xspace} 

\fancypagestyle{firstpage}{
  \fancyhf{}

  \pagenumbering{arabic}
}

\title{\OURLCORE\ Deep Learning Computing} 
\author{Sayeh Sharify, Mostafa Mahmoud, Alberto Delmas Lascorz, Milos Nikolic, Andreas Moshovos\\
Electrical and Computer Engineering, University of Toronto\\\{sayeh, delmasl1, moshovos\}@ece.utoronto.ca,\\ \{mostafa.mahmoud, milos.nikolic\}@mail.utoronto.ca} %

\begin{document}
\maketitle
\thispagestyle{firstpage}
\pagestyle{plain}

\begin{abstract}
We motivate a method for transparently identifying ineffectual computations in unmodified Deep Learning models and without affecting accuracy. Specifically, we show that if we decompose multiplications down to the bit level the amount of work performed during inference for image classification models can be consistently reduced by \textit{two orders of magnitude}. In the best case studied of a sparse variant of AlexNet, this approach can ideally reduce computation work by more than $500\times$. We present \OURL a hardware accelerator that implements this approach to improve execution time, and energy efficiency for inference with Deep Learning Networks. \OURL judiciously gives up some of the work reduction potential to yield a low-cost, simple, and energy efficient design that outperforms other state-of-the-art accelerators. For example, a \OURL configuration that uses a weight memory interface with just 128 wires outperforms a conventional accelerator with a 2K-wire weight memory interface by $2.3\times$ on average while being $2.13\times$ more energy efficient on average. A \OURL configuration that uses a 1K-wire weight memory interface, outperforms the 2K-wire conventional accelerator by $15.4\times$ and is $1.95\times$ more energy efficient. \OURL does not require but rewards advances in model design such as a reduction in precision, the use of alternate numeric representations that reduce the number of bits that are ``1'', or an increase in weight or activation sparsity.      

\end{abstract}

\section{Motivation}
\label{sec:intro}


Modern computing hardware is energy-constrained and thus developing techniques that reduce the amount of energy required to perform the computation is essential for improving performance. The bulk of the work performed by convolutional neural networks during inference is due to 2D convolutions (see Section~\ref{sec:bg}). In turn, these convolutions entail numerous multiply-accumulate operations were most of the work is due to the multiplication of an activation $A$ and a weight $W$. In order to improve energy efficiency a hardware accelerator can thus strive to perform only those multiplications that are effectual which will also lead to fewer additions. We can approach a $A\times W$ multiplication as a monolithic action which can be either performed or avoided in its entirety. Alternatively, we can decompose it into a collection of simpler operations. For example, if $A$ and $W$ are 16b fixed-point numbers $A\times W$ can be approached as 256 $1b\times 1b$ multiplications or 16 $16b\times 1b$ ones.

\begin{figure*}[!t]
\centering
\subfloat[]{
\centering
\includegraphics[width=0.8\textwidth]{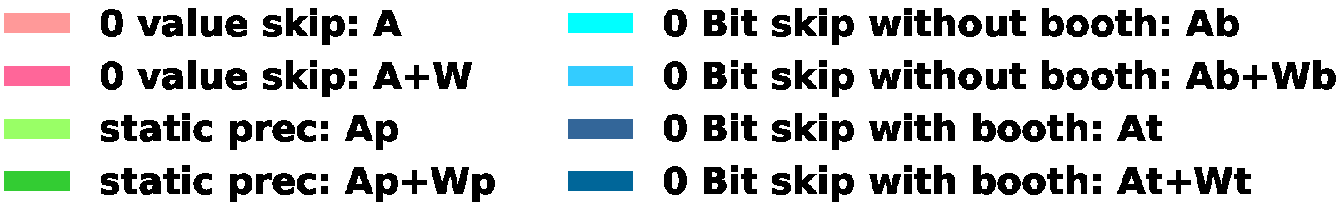}
\label{fig:legend}
}
\\
\subfloat[AlexNet]{
\centering
\includegraphics[width=0.33\textwidth]{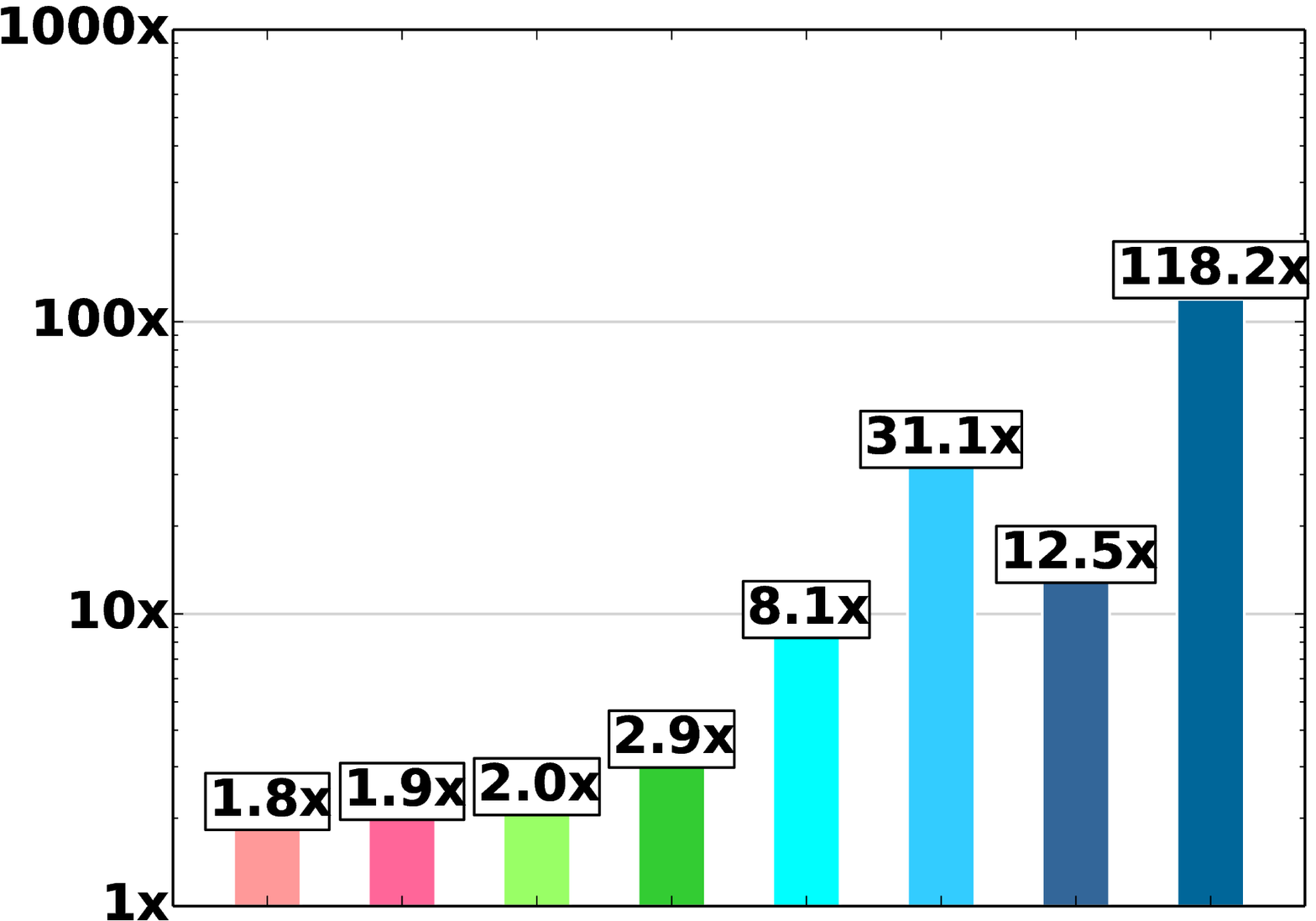}
\label{fig:alexnet_potential}
}
\subfloat[GoogLeNet]{
\centering
\includegraphics[width=0.33\textwidth]{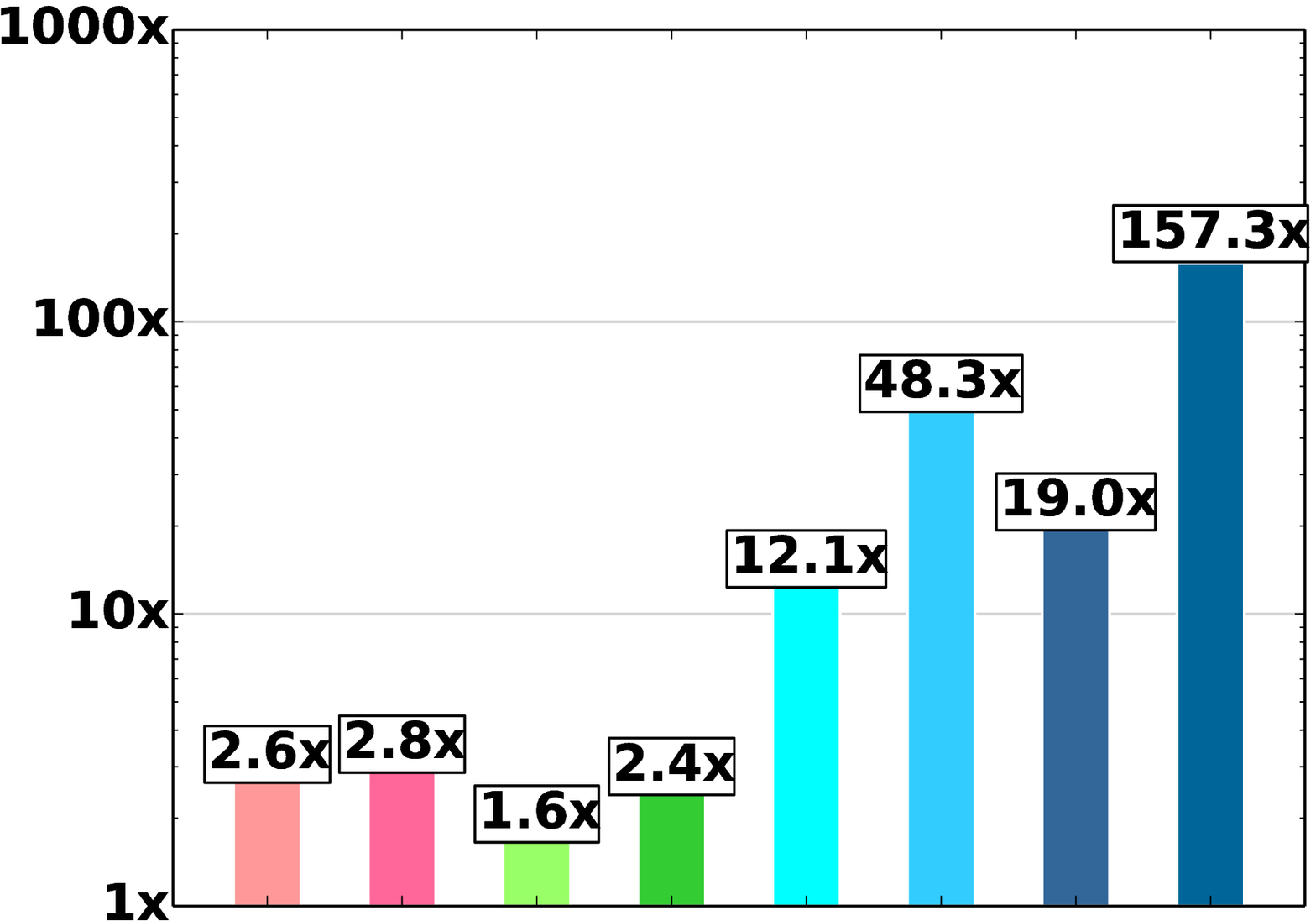}
\label{fig:google_potential}
}
\subfloat[VGG\_S]{
\centering
\includegraphics[width=0.33\textwidth]{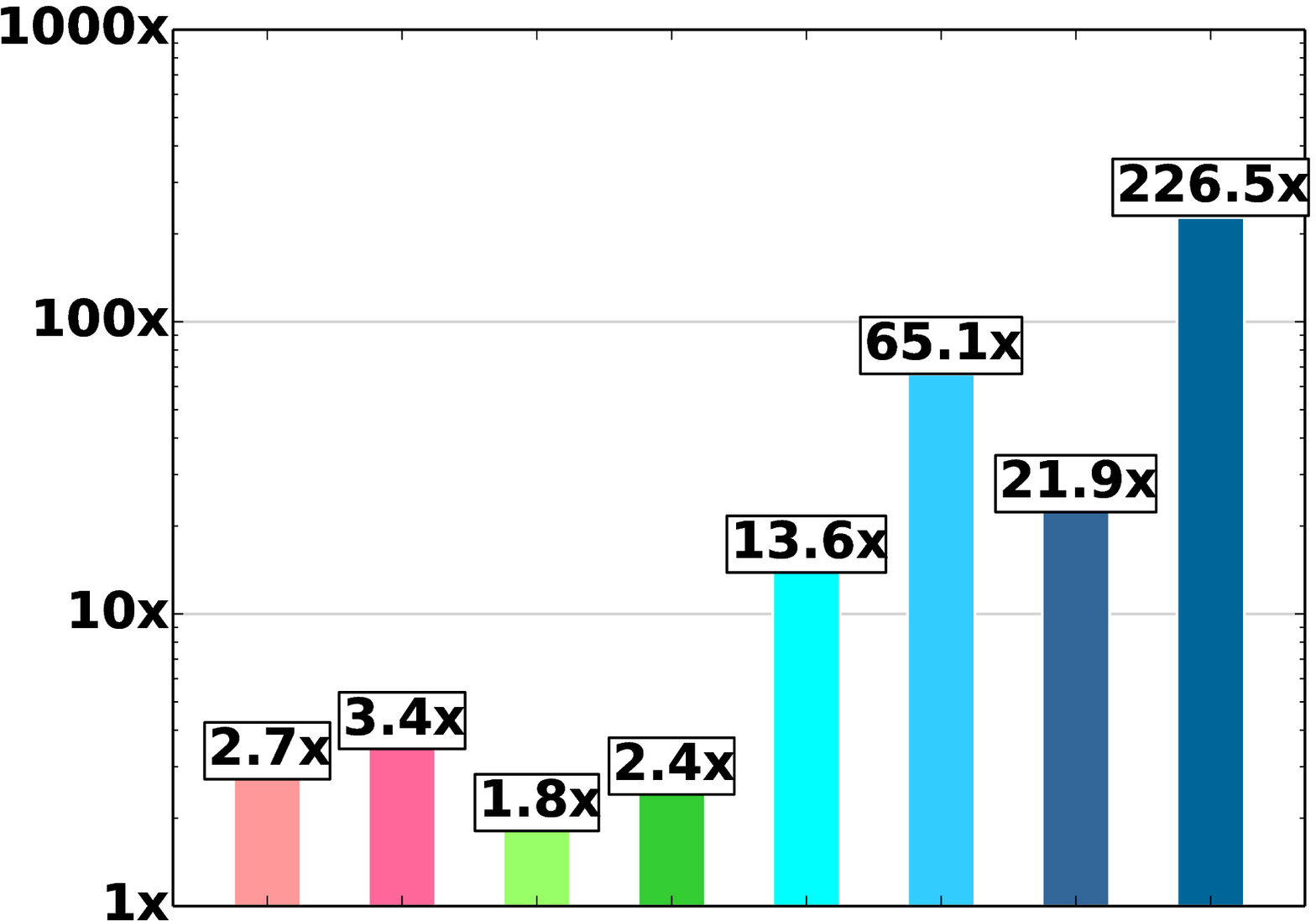}
\label{fig:vggs_potential}
}
\\
\subfloat[VGG\_M]{
\centering
\includegraphics[width=0.33\textwidth]{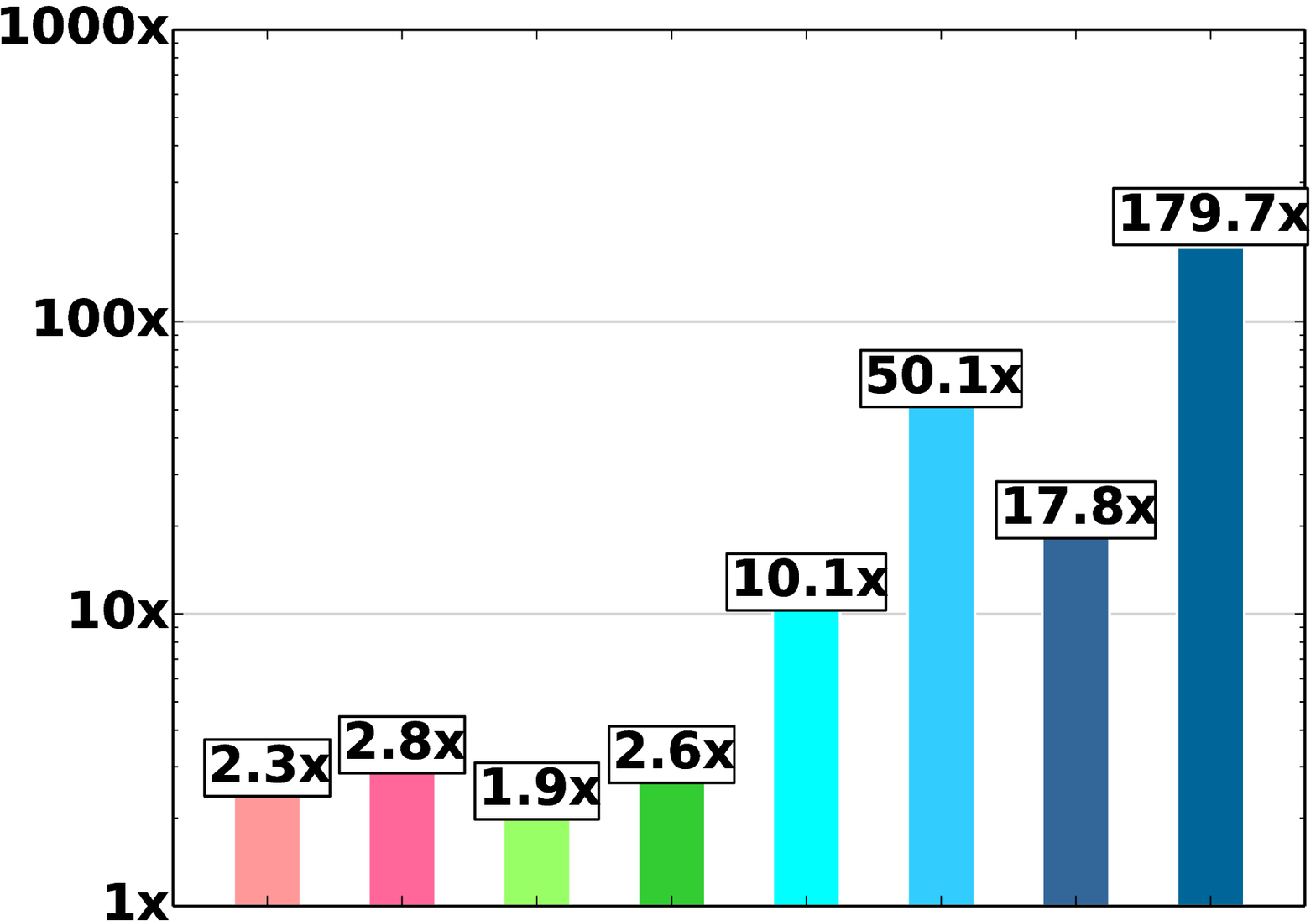}
\label{fig:vggm_potential}
}
\subfloat[AlexNet-Sparse]{
\centering
\includegraphics[width=0.33\textwidth]{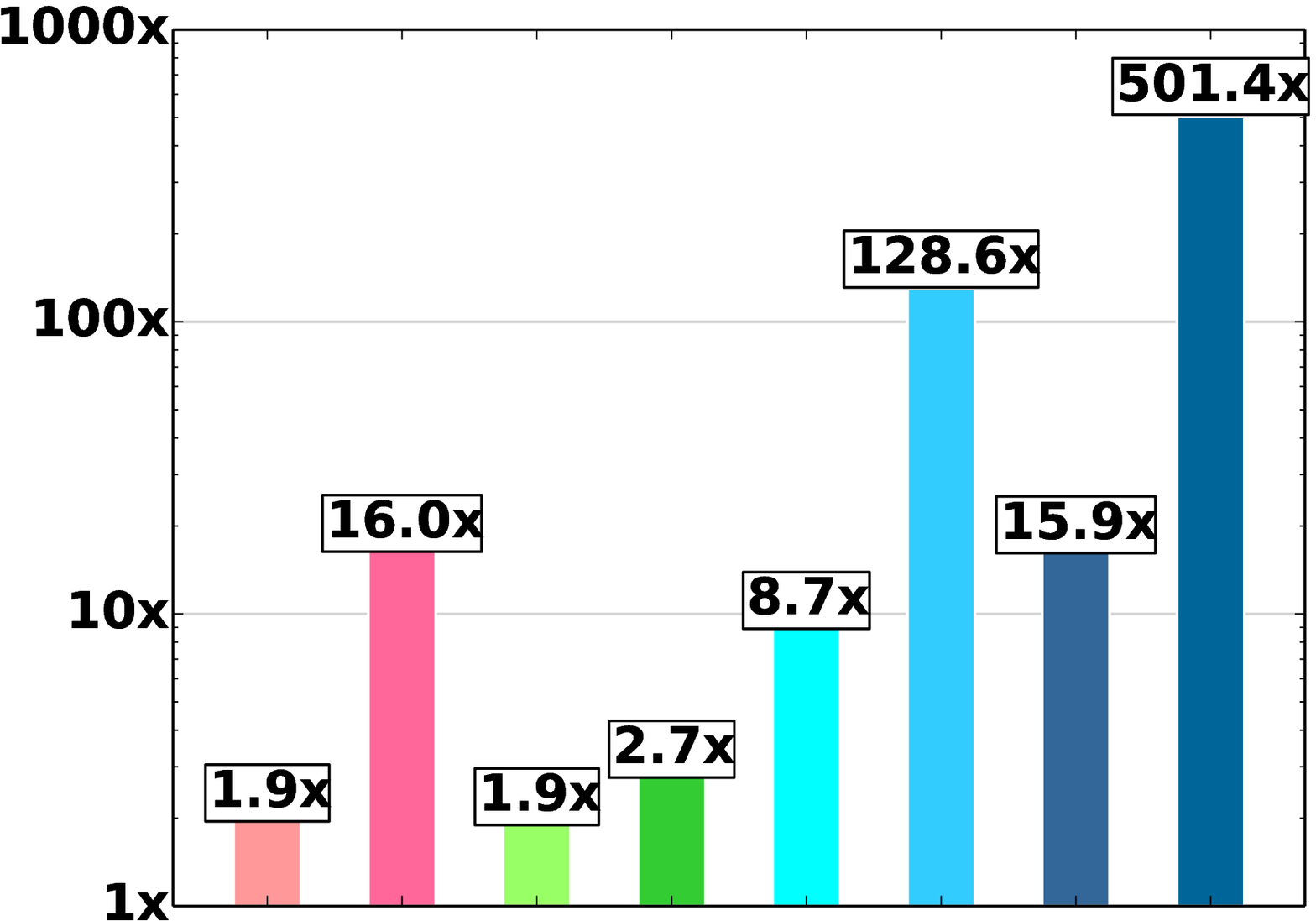}
\label{fig:alexnet-sparse_potential}
}
\subfloat[ResNet-Sparse]{
\centering
\includegraphics[width=0.33\textwidth]{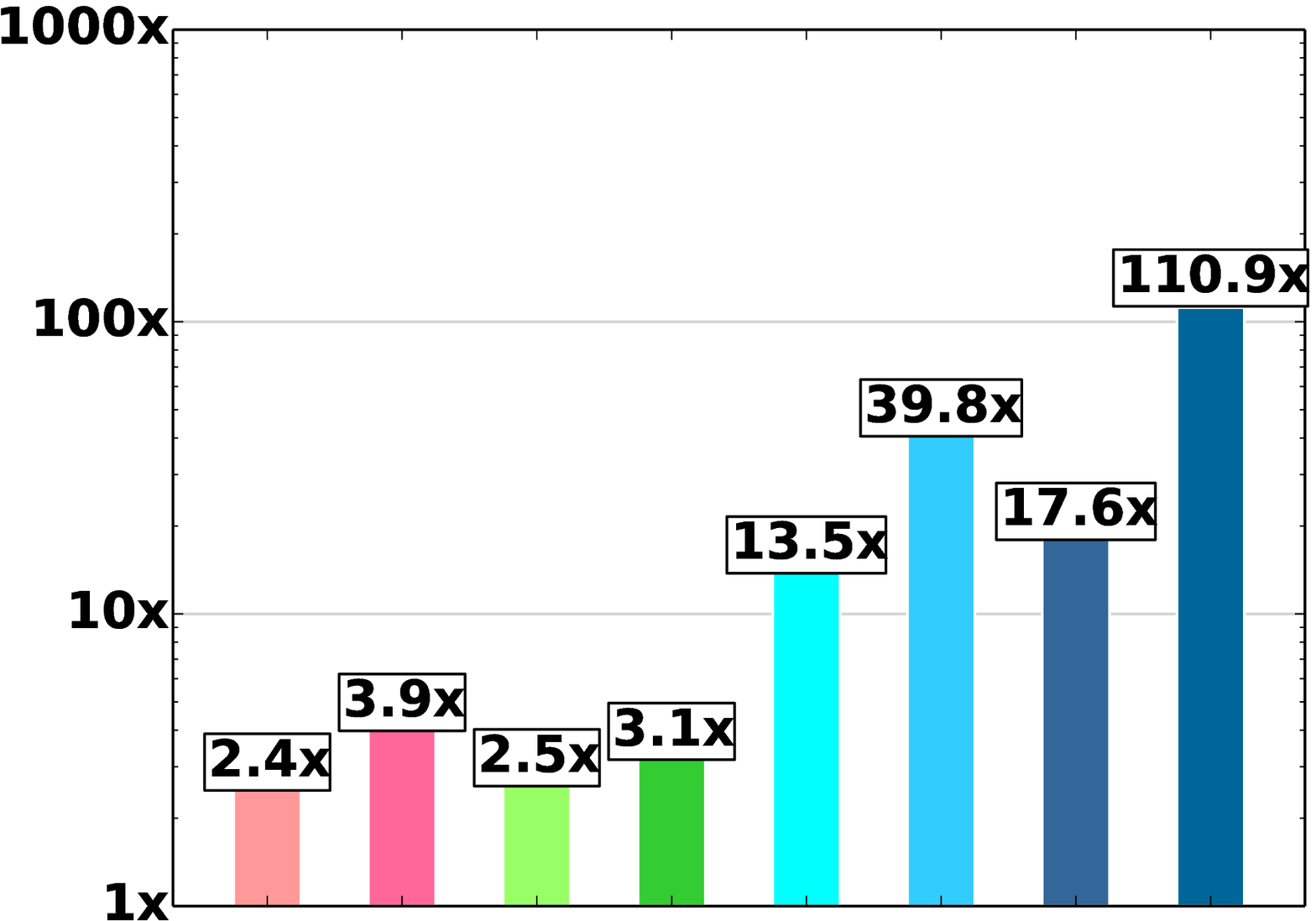}
\label{fig:resnet-sparse}
}
\\
\subfloat[Geomean]{
\centering
\includegraphics[width=0.33\textwidth]{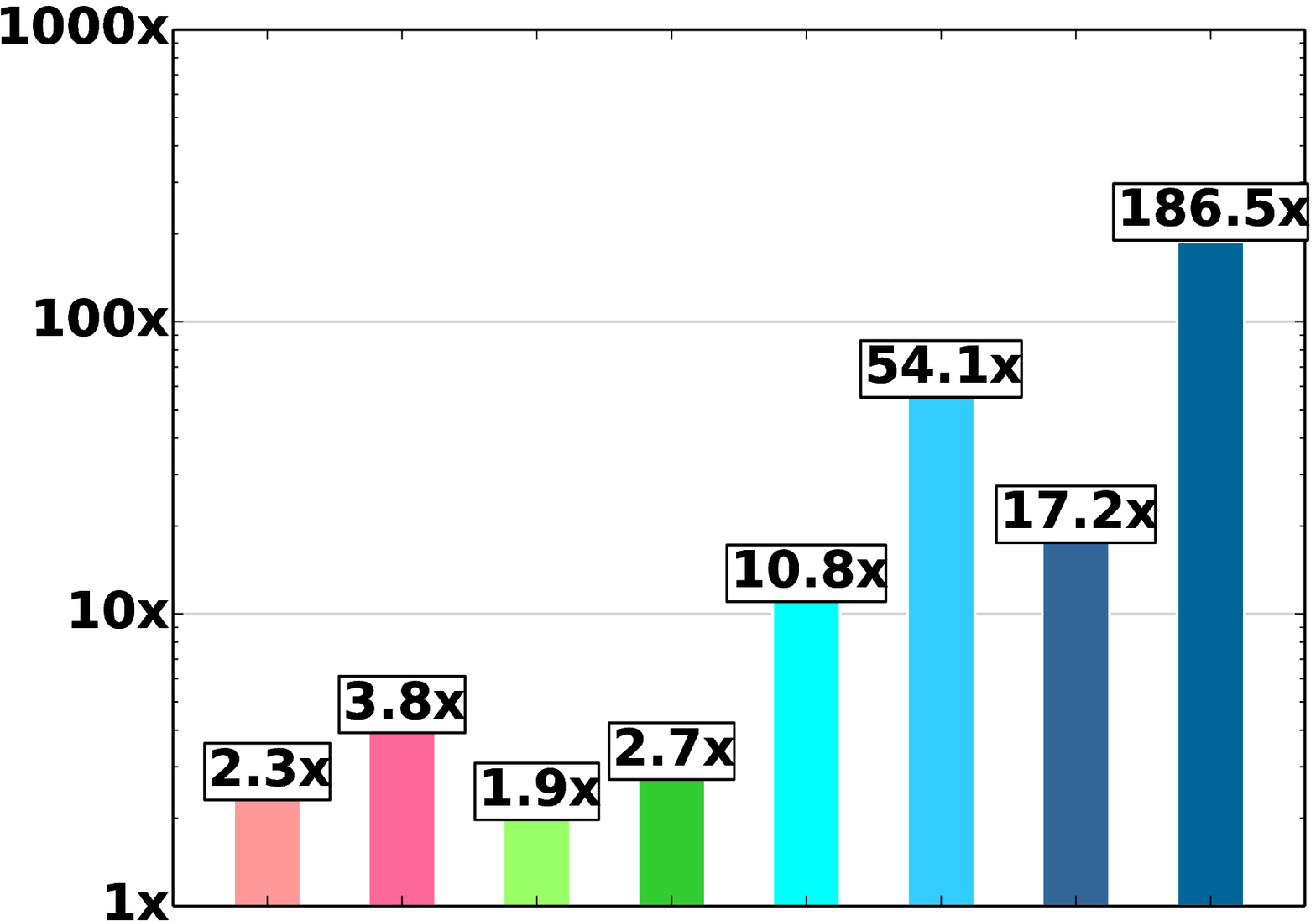}
\label{fig:geomean_potential}
}
\caption{Performance improvement potential for: 1) skipping zero activations~\cite{Albericio2016}, 2) skipping zero activations and weights, 3) using static precision for activations~\cite{Stripes-MICRO}, 4) using static precision for activations and weights, 5) skipping zero bits of activations~\cite{pragmatic}, 6) skipping zero bits of activations and weights, 7) skipping zero bits of activations using booth encoding~\cite{pragmatic}, 8) skipping zero bits of activations and weights using booth encoding (logarithmic scale).}
\label{fig:potential}
\end{figure*}

Figure~\ref{fig:geomean_potential} reports the potential reduction in work for several ineffectual work avoidance policies. The ``A'' policy avoids multiplications where the activation is zero. This is representative of the first generation of value-based accelerators that were motivated by the relatively large fraction of zero activations that occur in convolutional neural networks, e.g., Cnvlutin~\cite{Albericio2016}. The ``A+W'' skips those multiplications where either the activation or the weight are zero and is representative of accelerators that target sparse models where a significant fraction of synaptic connections has been pruned, e.g., SCNN~\cite{SCNN}. The ``Ap'' (e.g., Stripes~\cite{Stripes-MICRO} or Dynamic Stripes~\cite{dynamicstripes}) and ``Ap+Wp'' (e.g., Loom~\cite{Loom}) policies target precision for the activations alone or for the activations and the weights respectively. It has been found that neural networks exhibit variable per layer precision requirements. All aforementioned measurements corroborate past work on accelerator designs that exploited the respective properties.

However, we show that further potential for work reduction exists if we decompose the multiplications at the bit level. Specifically, for our discussion we can assume without loss of generality that these multiplications operate on 16b fixed-point values. The multiplication itself is given by:

\begin{equation}
A \times W = \sum_{i=0}^{15}\sum_{j=0}^{15}A_i\ \textsc{and}\ W_j
\end{equation}

where $A_i$ and $W_j$ are bits of $A$ and $W$ respectively.
When decomposed down to the individual 256 single bit multiplications one can observe that it is only those multiplications where both $A_i$ and $W_j$ are non-zero that are \textit{effectual}. Accordingly, the ``Ab'' (e.g., Pragmatic~\cite{pragmatic}) and ``Ab+Wb'' measurements show the potential reduction in work that is possible if we skip those single bit multiplications where the activation bit is zero or whether either the activation or the weight bits are zero respectively. The results show that the potential is far greater than the policies discussed thus far. 

Further to our discussion, rather than representing $A$ and $W$ as bit vectors, we can instead Booth-encode them as a series of signed powers of two, or \textit{terms} (higher-radix Booth encoding is also possible). In this case the multiplication is given by:

\begin{equation}
A \times W = \sum_{i=0}^{A_{terms}}\sum_{j=0}^{W_{terms}}{At_i}\ \times\ {Wt_j}
\end{equation}

where $At_i$ and $Wt_i$ are of the form $\pm2^x$. As with the positional representation, it is only those products where both $At_i$ and $Wt_i$ are non-zero that are effectual. Accordingly, the figure shows the potential reduction in work with ``At'' where we skip the ineffectual terms for a Booth-encoded activation (e.g., Pragmatic~\cite{pragmatic}), and with ``At+Wt'' where we calculate only those products where both the activation and the weight terms are non-zero. The results show that the reduction in work (and equivalently the performance improvement potential) with ``At+Wt''  is in most cases two orders of magnitude higher than the zero value or the precision based approaches.

Based on these results, our goal is to develop a hardware accelerator that computes only the effectual terms. No other accelerator to date has exploited this potential. Moreover, by targeting ``At+Wt'' we can also exploit ``Ab+Wb'' where the inputs are represented in a plain positional representation and are not Booth-encoded.

\section{Background}
\label{sec:bg}
This section provides the required background information as follows: Section~\ref{sec:cvl} reviews operation of a Convolutional Neural Network  and Section~\ref{sec:base} goes through our baseline system.

\begin{figure}[t]
\includegraphics[scale=0.5]{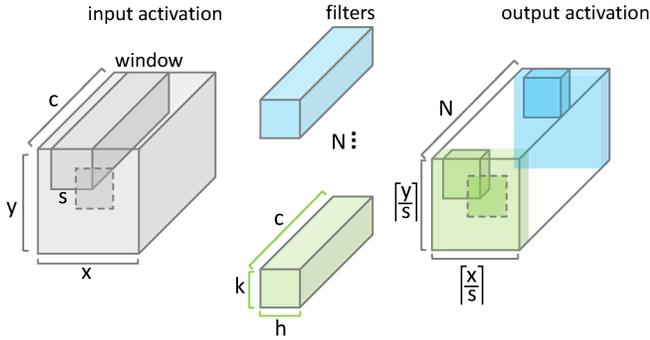}
\caption{Convolutional Layer}
\label{fig:conv}
\end{figure}

\subsection{Convolutional Layers}
\label{sec:cvl}

Convolutional Neural Networks (CNNs) usually consist of several Convolutional layers (CVLs) followed by a few fully-connected layers (FCLs). In many image related CNNs most of the operation time is spent on processing CVLs in which a 3D convolution operation is applied to the input activations producing output activations. Figure~\ref{fig:conv} illustrates a CVL with a $c\times x\times y$ input activation block and $N$ $c\times h\times k$ filters. The layer dot products each of these $N$ filters (denoted $f^0, f^1, ..., f^{N-1}$) by a $c\times h\times k$ subarray of input activation, called \textit{window}, to generate a single $o_h \times o_k$ output activation. In total convolving $N$ filters and an activation window results in $N$ $o_h \times o_k$ outputs which will be passed to the input of the next layer. The convolution of activation windows and filters takes place in a sliding window fashion with a constant stride $S$.

Fully-connected layers can be implemented as convolutional layers in which filters and input activations have the same dimensions, i.e., $x=h$ and $y=k$.

\begin{figure*}[t]
\includegraphics[scale=1.23]{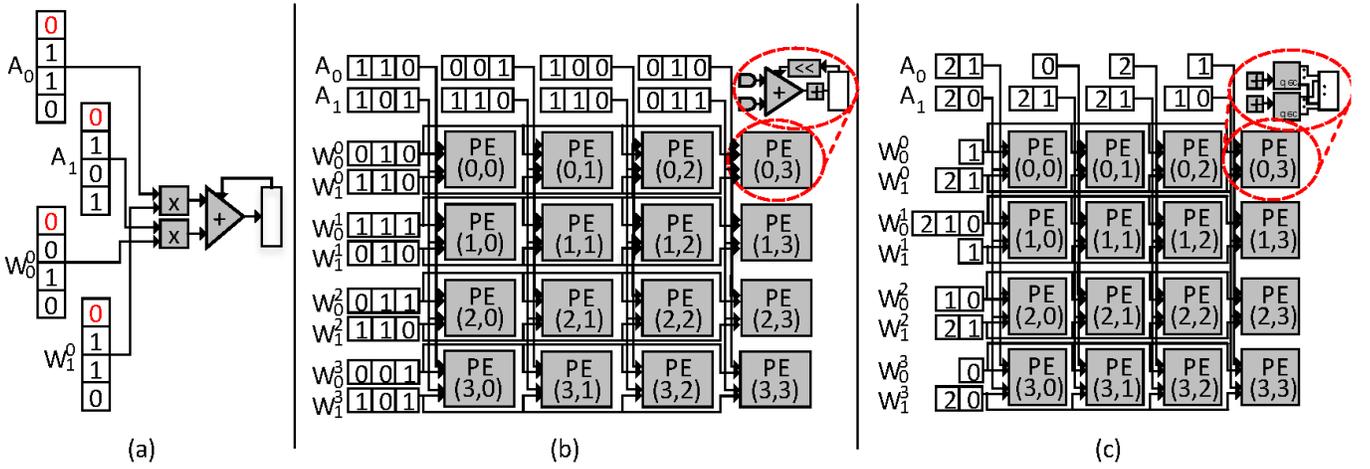}
\caption{a) Bit-parallel unit. b) Bit-serial unit with equivalent throughput~\cite{Loom}. c) \OURL  unit with equivalent throughput where for both activations and weights only essential information is processed.}
\label{fig:example}
\end{figure*}

\subsection{Baseline system}
\label{sec:base}
Our baseline design (\BASE) is a data-parallel engine inspired by the DaDianNao accelerator~\cite{DaDiannao} which uses 16-bit fixed-point activations and weights. Our baseline configuration has 8 inner product units (IPs) each accepting 16 input activations and 16 weights as inputs. The 16 input activations are broadcast to all 8 IPs; however, each IP has its own 16 weights. Every cycle each IP multiplies 16 input activations by their 16 corresponding weights and reduces them into a single partial output activation using a 16 32-bit input adder tree. The partial results are accumulated over the multiple cycles to generate the final output activation. An activation memory provides the activations and a weight memory provides the weights. Other memory configurations are possible.

\section{\OURLCORE: a simplified example} 
This section illustrates the key concepts behind \OURL via an example using 4-bit activations and weights.

\noindent\textbf{Bit-Parallel Processing: } Figure~\ref{fig:example}a shows a bit-parallel engine multiplying two 4-bit activation and weight pairs, generating a single 4-bit output activation per cycle. Its throughput is two $4b\times 4b$ products per cycle.  

\noindent\textbf{Bit-Serial Processing: } Figure ~\ref{fig:example}b shows an equivalent bit-serial engine which is representative of Loom (\BSD)~\cite{Loom}. To match the bit-parallel engine's throughput, \BSD processes 8 input activations and 8 weights every cycle producing 32 $1b \times 1b$ products. Since \BSD processes both activations and weights bit-serially, it produces 16 output activations in $P_a \times P_w$ cycles where $P_a$ and $P_w$ are the activation and weight precisions, respectively. Thus, \BSD outperforms the bit-parallel engine by $\frac{16}{P_a\times P_w}$. In this example, since both activations and weights can be represented in three bits, the speedup of \BSD over the bit-parallel engine is $1.78\times$. However, \BSD still processes some ineffectual terms. For example, in the first cycle 27 of the 32 $1b\times1b$ products are zero and thus ineffectual and can be removed.

\noindent\textbf{\OURL:} Figure ~\ref{fig:example}c illustrates a simplified \OURL engine in which both the activations and weights are represented as vectors of essential powers of two, or \textit{one-offsets}. For example, $A_0$ = $(110)$ is represented as a vector of its one-offsets $A_0$ = $(2,1)$. Every cycle each PE accepts a 4-bit one-offset of an input activation and a 4-bit one-offset of a weight and adds them up to produce the power of the corresponding product term in the output activation. Since \OURL processes activation and weight ``term''-serially, it takes $t_a\times t_w$ cycles for each PE to complete producing the product terms of an output activation, where $t_a$ and $t_w$ are the number of one-offsets in the corresponding input activation and weight. The engine processes the next set of activation and weight one-offsets after $T$ cycles, where $T$ is the maximum $t_a\times t_w$ among all the PEs. In this example, the maximum $T$ is 6 corresponding to the pair of $A_0=(2,1)$ and ${W_0}^1=(2,1,0)$ from $PE(1,0)$. Thus, the engine can start processing the next set of activations and weights after 6 cycles achieving $2.67\times$ speedup over the bit-parallel engine.

\section{\OURLCORE}
\label{sec:ea}
This section presents the \OURL architecture by explaining its processing approach, processing elements structure, and its high-level organization. 

\subsection{Approach}
\OURL's goal is to minimize the required computation for producing the products of input activations and weights by processing only the essential bits of both the input activations and weights. To do so, \OURS converts, on-the-fly, the input activations and weights into a representation which contains only the essential bits, and processes per cycle one pair of essential bits one from an activation and another from a weight. The rest of this section is organized as follows: Section~\ref{sec:rep} describes the activation and weight representations in \OURS and Section~\ref{sec:term} explains how \OURS calculates the product terms.

\begin{figure*}[h]
\subfloat[]{
\centering
\includegraphics[width=0.74\textwidth]{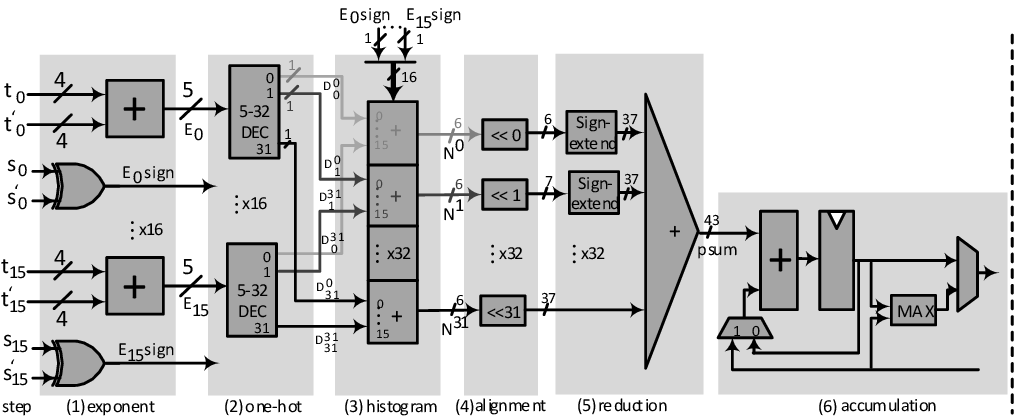}
\label{fig:PEa}
}
\subfloat[]{
\centering
\includegraphics[width=0.25\textwidth]{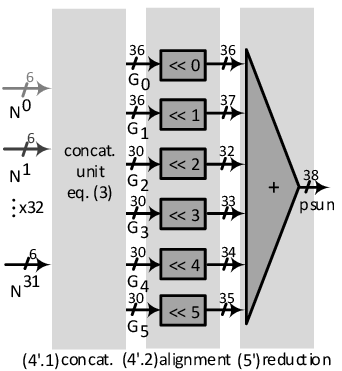}
\label{fig:PEb}
}
\caption{\OURL processing element, a) 1) Calculating the exponents of the products, 2) Converting the exponents into their corresponding one-hot format, 3) Counting the number of each power of two bucket, 4) Shifting the counting results according to the bucket values, 5) Sign-extending the shifted values and reducing the results into a single 42-bit partial output, 6) Accumulating the partial outputs over multiple cycles. b) Enhanced steps (4) and (5).}
\label{fig:PE}
\end{figure*}

\subsubsection{Activation and Weight Representation}
\label{sec:rep}
For clarity we present a \OURS implementation that processes the one-offsets, that is the non-zero signed powers of two in a Booth-encoded representation of the activations and weights (however, \OURS could be adjusted to process a regular positional representation or adapted to process representations other than fixed-point). \OURS represent each activation or weight as a list $(o_n,\ldots,o_0)$ of its one-offsets. Each one-offset is represented as $(sign, magnitude)$ pair. For example, an activation $A=\text{-}2_{(10)} = 1110_{(2)}$ with a Booth-encoding of $0010_{(2)}$ would be represented as $(\text{-},1)$ and a $A=7_{(10)} = 0111_{(2)}$ will be presented as $((\text{+},3),(\text{-},0))$. The sign can be encoded using a single bit, with, for example, $0$ representing ``$\text{+}$'' and $1$ representing ``$\text{-}$''.


\subsubsection{Calculating a Product Term}
\label{sec:term}
\sloppy
\OURS calculates the product of a weight $W=(W_{terms})$ and an input activation $A=(A_{terms})$ where each term is a $(sign,magnitude)=(s_i, t_i)$ as follows:

\begin{footnotesize}
\begin{equation} \label{eq:calc}
\footnotesize
\begin{aligned}
W\times A &=  \sum_{\forall (s,t) \in W{terms}}  (\text{-}1)^{s} 2^{t} \times \sum_{\forall (s^\prime, t^{\prime}) \in A{terms}} (\text{-}1)^{s^\prime}2^{t^{\prime}}\\
&= ((\text{-}1)^{s_0}2^{t_0}+(\text{-}1)^{s_1}2^{t_1}+\cdots+(\text{-}1)^{s_n}2^{t_n})\times\\ 
&\ \ \ \ \ \ ((\text{-}1)^{s^{\prime}_0}2^{t^{\prime}_0}+(\text{-}1)^{s^{\prime}_1}2^{t^{\prime}_1}+\cdots+(\text{-}1)^{s^{\prime}_m}2^{t^{\prime}_m})
\\\\
&= ((\text{-}1)^{s_0}(\text{-}1)^{s^{\prime}_0}(2^{t_0}\times2^{t^{\prime}_0})+\cdots+((\text{-}1)^{s_0}(\text{-}1)^{s^{\prime}_m}2^{t_0}\times2^{t^{\prime}_m}))+\\
&\ \ \ \ \ \ \cdots+((\text{-}1)^{s_n}(\text{-}1)^{s^{\prime}_0}(2^{t_n}\times2^{t^{\prime}_0}) +\cdots+((\text{-}1)^{s_n}(\text{-}1)^{s^{\prime}_m}2^{t_n}\times2^{t^{\prime}_m}))
\\\\
&= ((\text{-}1)^{(s_0+s^{\prime}_0)}2^{(t_0+t^{\prime}_0)}+\cdots+(\text{-}1)^{(s_0+s^{\prime}_m)}2^{(t_0+t^{\prime}_m)})+\\
&\ \ \ \ \ \ \cdots+((\text{-}1)^{(s_n+s^{\prime}_0)}2^{(t_n+t^{\prime}_0)}+\cdots+(\text{-}1)^{(s_n+s^{\prime}_m)}2^{(t_n+t^{\prime}_m)})
\end{aligned}
\end{equation}
\end{footnotesize}

That is, instead of processing the full $A\times W$ product in a single cycle, \OURS processes each product of a single $t^{\prime}$ term of the input activation $A$ and of a single $t$ term of the weight $W$ individually. Since these terms are powers of two so will be their product. Accordingly, \OURS can first add the corresponding exponents $t^{\prime}+t$. If a single product is processed per cycle, the $2^{t^{\prime}+t}$ final value can be calculated via a decoder. In the more likely configuration where more than one term pairs are processed per cycle, \OURS can use one decoder per term pair to calculate the individual $2^{t^{\prime}+t}$ products and then an efficient adder tree to accumulate all. This is described in more detail in the next section. 



\subsection{Processing Element}
Figure~\ref{fig:PE} illustrates how the \OURS Processing Element (PE) calculates the product of a set of weights and their corresponding input activations. Without loss of generality we assume that each PE multiplies $16$ weights, $W_0$,...,$W_{15}$, by 16 input activations, $A_0$,...,$A_{15}$. The PE calculates the 16 products in 6 steps: 

In \textbf{Step 1}, the PE accepts 16 4-bit weight one-offsets, $t_0$,\ldots,$t_{15}$ and their 16 corresponding sign bits $s_0$,\ldots,$s_{15}$, along with 16 4-bit activation one-offsets, $t^{\prime}_0$,\ldots,$t^{\prime}_{15}$ and their signs $s^{\prime}_0$,\ldots,$s^{\prime}_{15}$, and calculates 16 one-offset pair products. Since all one-offsets are powers of two, their products will also be powers of two. Accordingly, to multiply 16 activations by their corresponding weights \OURS adds their one-offsets to generate the 5-bit exponents ($t_0+t^{\prime}_0$),\ldots,($t_{15}+t^{\prime}_{15}$) and uses 16 XOR gates to determine the signs of the products. 

In \textbf{Step 2}, for the $i^{th}$ pair of activation and weight, where $i$ is $\in$\{0,...,15\}, the PE calculates $2^{(t_i+t^{\prime}_i)}$ via a 5b-to-32b decoder which converts the 5-bit exponent result ($t_i+t^{\prime}_i$) into its corresponding one-hot format, i.e., a 32-bit number with one ``1'' bit and 31 ``0'' bits. The single ``1'' bit in the $j^{th}$ position of a decoder output corresponds to a value of either $\text{+}2^j$ or $\text{-}2^j$ depending on the sign of the corresponding product ($E_i.sign$ on the figure). 

\textbf{Step 3: } The PE generates the equivalent of a histogram of the decoder output values. Specifically, the PE accumulates the 16 32-bit numbers from Step 2 into $32$ buckets, $N^0,\ldots,N^{31}$ corresponding to the values of $2^0, 2^1, ..., 2^{31}$ as there are $32$ powers of two. The signs of these numbers $E_i.sign$ from Step 1 are also taken into account. At the end of this step, each ``bucket'' contains the count of the number of inputs that had the corresponding value.  Since each bucket has $16$ signed inputs the resulting count would be in a value in $[\text{-}16,...,16]$ and thus is represented by $6$ bits in 2's complement.

\textbf{Step 4: }Na\"ively reducing the 32 6-bit counts into the final output would require first ``shifting'' the counts according to their weight converting all to $31+6=37$b and then using a 32-input adder tree as shown in Figure~\ref{fig:PEa}(4)-(5). Instead \OURS reduces costs and energy by exploiting the relative weighting of each count by grouping and concatenating them in this stage as shown in Figure~\ref{fig:PEb}($4^{\prime}.1$).  For example, rather than adding $N^0$ and $N^6$ we can simply concatenate them as they are guaranteed to have no overlapping bits that are ``1''.
This is explained in more detail in Section~\ref{sec:adderTree}.

\textbf{Step 5: } As Section~\ref{sec:adderTree} explains in more detail, the concatenated values from Step $4^{\prime}.1$ are added via a 6-input adder tree as shown in Figure~\ref{fig:PEb}($5^{\prime}$) producing a $38$b partial sum.

\textbf{Step 6: } The partial sum from the previous step is accumulated with the partial sum held in an accumulator. This way, the complete $A\times W$ product can be calculated over multiple cycles, one effectual pair of one-offsets per cycle.


The aforementioned steps are not meant to be interpreted as pipeline stages. They can be merged or split as desired.

\subsubsection{Enhanced Adder Tree}
\label{sec:adderTree}
\begin{figure}[t]
\includegraphics[scale=1.6]{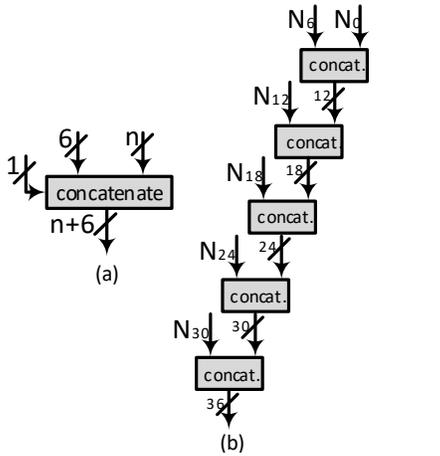}
\centering
\caption{One of \OURL's concatenation units.}
\label{fig:concat}
\end{figure}

Step 5 of Figure~\ref{fig:PEa} has to add $32$ $6$b counts each weights by the corresponding power of 2. This section presents an alternate design that replaces Steps 4 and 5. Specifically, it presents an equivalent more area and energy efficient ``adder tree'' which takes advantage of the fact that the outputs of Step 4 contain groups of numbers that have no overlapping bits that are ``1''. For example, in relation to the na\"ive adder tree of Figure~\ref{fig:PEa}(5) consider adding the $6^{th}$ 6-bit input 
($N^\textbf{6}\text{=}n^{\textbf{6}}_5 n^{\textbf{6}}_4 n^{\textbf{6}}_3 n^{\textbf{6}}_2 n^{\textbf{6}}_1 n^{\textbf{6}}_0$) 
with the $0^{th}$ 6-bit input 
($N^\textbf{0}\text{=}n^{\textbf{0}}_5 n^{\textbf{0}}_4 n^{\textbf{0}}_3 n^{\textbf{0}}_2 n^{\textbf{0}}_1 n^{\textbf{0}}_0$).
We have to first shift $N^6$ by $6$ bits which amounts to adding $6$ zeros as the $6$ least significant bits of the result. In this case, there will be no bit position in which both $N^6\text{<<}\ 6$ and $N^0$ will have a bit that is 1. Accordingly, adding ($N^6 \text{<<}\ 6$) and $N^0$ is equivalent to concatenating either $N^6$ and $N^0$ or ($N^6$-1) and $N^0$ based on the sign bit of $N^0$ (Figure~\ref{fig:concat}a):

\begin{small}
\begin{equation} 
\begin{array}{l}
N^{6} \times 2^{6} + N^0 = (N^{6} << 6) + N^0
\\\\
\text{1) if $n^{0}_5$ is zero:}
\\\\
\hspace*{7mm}= n^{6}_5 n^{6}_4 n^{6}_3 n^{6}_2 n^{6}_1 n^{6}_0 000000+000000 n^{0}_5 n^{0}_4 n^{0}_3 n^{0}_2 n^{0}_1 n^{0}_0
\\\\
\hspace*{7mm}= n^{6}_5 n^{6}_4 n^{6}_3 n^{6}_2 n^{6}_1 n^{6}_0 n^{0}_5 n^{0}_4 n^{0}_3 n^{0}_2 n^{0}_1 n^{0}_0 = \{N^6,N^0\}
\\\\
\text{2) else if $n^{0}_5$ is one:}
\\\\
\hspace*{7mm}= n^{6}_5 n^{6}_4 n^{6}_3 n^{6}_2 n^{6}_1 n^{6}_0 000000+111111 n^{0}_5 n^{0}_4 n^{0}_3 n^{0}_2 n^{0}_1 n^{0}_0
\\\\
\hspace*{7mm}= (n^{6}_5+1) (n^{6}_4+1) (n^{6}_3+1) (n^{6}_2+1) (n^{6}_1+1) (n^{6}_0+1) n^{0}_5 n^{0}_4 n^{0}_3 n^{0}_2 n^{0}_1 n^{0}_0
\\\\
\hspace*{7mm}= \{(N^6-1),N^0\}
\end{array}
\end{equation}
\end{small}
Accordingly, this process can be applied recursively, by grouping those $N^i$ where $(i\ \textsc{mod}\ 6)$ is equal. That is the $i^{th}$ input would be concatenated with $(i+6)^{th}$, $(i+12)^{th}$, and so on.  Figure~\ref{fig:concat}b shows an example unit for those $N^i$ inputs where $(i\ \textsc{mod}\ 6) = 0$. While the figure shows the concatenation done as stack, other arrangements are possible. 

For the 16 product unit described here the above process yields the following six groups:

\begin{small}
\begin{equation} 
\begin{array}{l}

 G_0=\{N^{30},N^{24},N^{18},N^{12},N^{6},N^{0}\}
\\\\
G_1=\{N^{31},N^{25},N^{19},N^{13},N^{7},N^{1}\}
\\\\
G_2=\{N^{26},N^{20},N^{14},N^{8},N^{2}\}
\\\\
G_3=\{N^{27},N^{21},N^{15},N^{9},N^{3}\}
\\\\
G_4=\{N^{28},N^{22},N^{16},N^{10},N^{4}\}
\\\\
G_5=\{N^{29},N^{23},N^{17},N^{11},N^{5}\}

\end{array}
\end{equation}
\end{small}

The final partial sum is  then given by the following:

\begin{small}
\begin{equation} 
psum = \sum_{i=0}^{5} (G_i << i)
\end{equation}
\end{small}

\begin{figure}[t]
\includegraphics[scale=0.44]{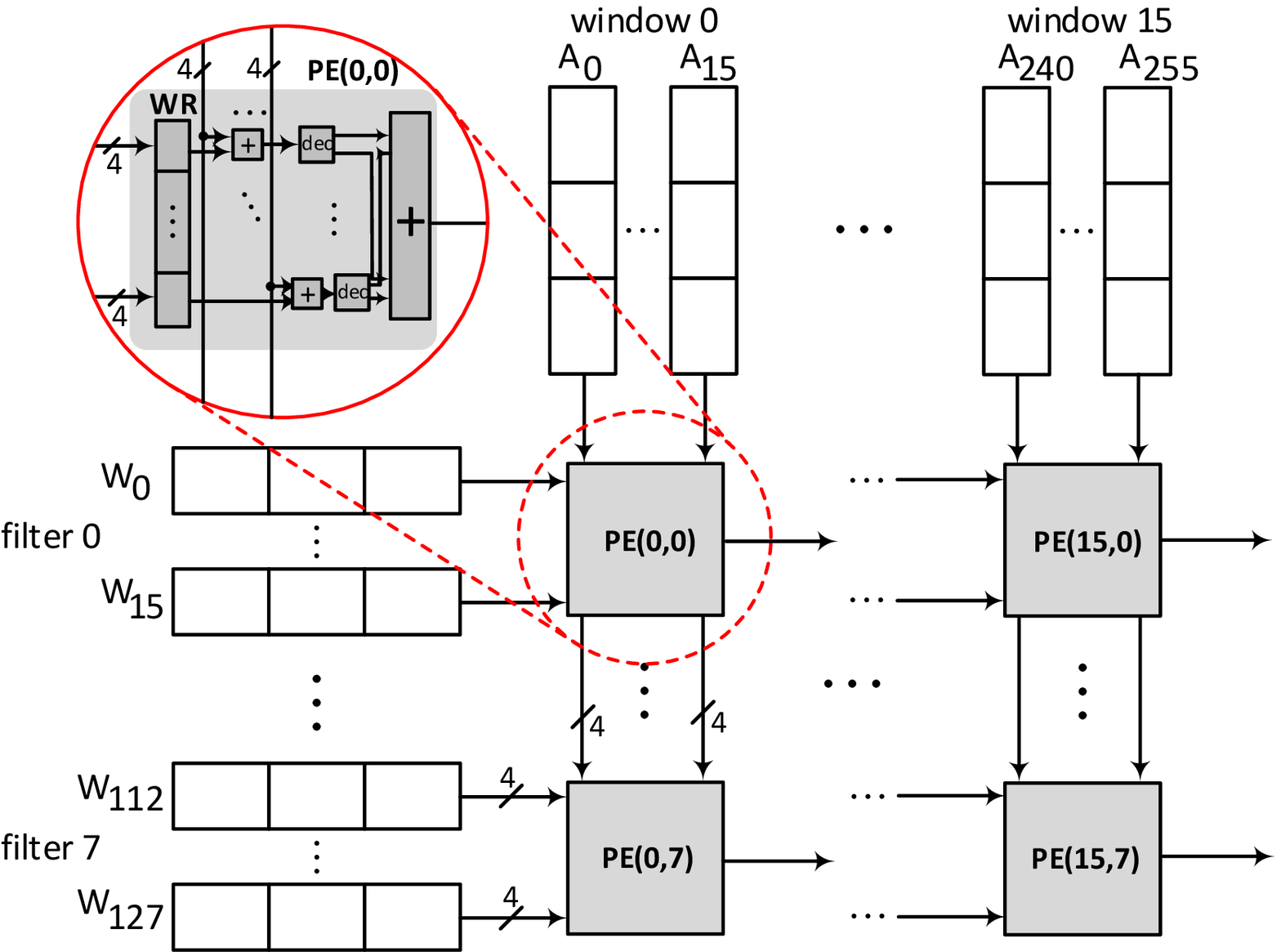}
\centering
\caption{\OURL tile}
\label{fig:LACtile}
\end{figure}

\subsection{Tile Organization}

Figure~\ref{fig:LACtile} illustrates \OURS tile which comprises a 2D array of PEs processing 16 windows of input activations and $K=8$ filters every cycle. PEs along the same column share the same input activations and PEs along the same row receive the same weights. Every cycle PE(i,j) receives the next one-offset from each input activation from the $j^{th}$ window and multiplies it by a one-offset of the corresponding weight from the $i^{th}$ filter. 
The tile starts processing the next set of activations and weights when all the PEs are finished with processing the terms of the current set of 16 activations and their corresponding weights. 

Since \OURS processes both activations and weights term-serially, to match our \BASE configuration it requires to process more filters or more windows concurrently. Here we consider implementations that process more filters. In the worst case each activation and weight possesses 16 terms, thus \OURS tile should process $8 \times 16 = 128$ filters in parallel to always match the peak compute bandwidth of \BASE. However, as shown in Figure~\ref{fig:potential} with $16\times$ more filters, \OURS's potential performance improvement over the baseline is more than two orders of magnitude. Thus, we can trade-off some of this potential by using fewer filters. 

\begin{figure}
\includegraphics[scale=1.9]{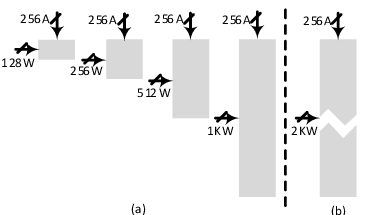}
\centering
\caption{a) \OURL configurations: $\OURS_{128}$, $\OURS_{256}$, $\OURS_{512}$, $\OURS_{1K}$. b) \BASE configuration: $\BASE_{2K}$}
\label{fig:config}
\end{figure}

To read weights from the WM \BASE requires 16 wires per weight while \OURS requires only one wire per weight as it process weights term-serially. Thus, with the same number of filters \OURS requires $16\times$ less wires. In this study we limit our attention to a \BASE configuration with 8 filters, 16 weights per filter, and thus $2K$ weight wires ($\BASE_{2K}$), and to \OURS configurations with 8, 16, 32, and 64 filters, and $128$, $256$, $512$, and $1K$ weight wires. In all designs, the number of activation wires is set to $256$ (Figure~\ref{fig:config}). Alternatively, we could fix the number of filters and accordingly number of weight wires and add more parallelism to the design by increasing the number of activation windows. The evaluation of such a design is not reported in this document.

\section{Evaluation}
\label{sec:eval}
This section evaluates \OURS's performance, energy and area and explores different configurations of \OURS comparing to $\BASE_{2K}$. This section considers $\OURS_{128}$, $\OURS_{256}$, $\OURS_{512}$, and $\OURS_{1k}$ configurations which require $128$, $256$, $512$, and $1K$ weight wires, respectively (Figure~\ref{fig:config}). 

\subsection{Methodology}
Execution time is modeled via a custom cycle-accurate simulator and energy and area results are reported based on post layout simulations of the designs. Synopsys Design Compiler~\cite{synopsys_site} was used to synthesize the designs with TSMC 65nm library. Layouts were produced with Cadence Innovus~\cite{Cadence_site} using synthesis results. Intel PSG ModelSim is used to generate data-driven activity factors to report the power numbers. The clock frequency of all designs is set to 1GHz. The ABin and ABout SRAM buffers were modeled with CACTI~\cite{Muralimanohar_cacti6.0:} and AM and WM were modeled as eDRAM with Destiny~\cite{destiny}.

\subsection{Performance}
\label{sec:e:p}
\begin{figure}
\includegraphics[scale=0.53]{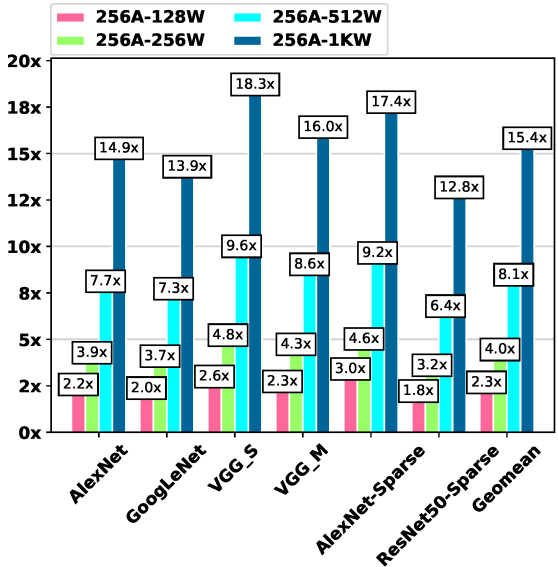}
\caption{\OURL performance relative to $\BASE_{2K}$}
\label{fig:performance}
\end{figure}

\begin{table*}
    \centering
    \footnotesize
    \caption{
Activation and weight precision profiles in bits for the convolutional layers. 
}
	\begin{tabular}{|l|p{5cm}|c|} 
\hline
                   & \multicolumn{2}{c|}{\textbf{Convolutional Layers}}                                                                                                                                                           \\ 
\cline{2-2}\cline{3-3}
 \textbf{Network}  & \multicolumn{2}{c|}{\textbf{100\% Accuracy} }                                                                                                                  \\ 
\cline{2-2}\cline{3-3}
                   & {\textbf{Activation Precision Per Layer} }                         & {\textbf{Weight Precision Per Network}} \\
\hline
AlexNet            & {9-8-5-5-7}                                       & {11}                   \\ 
\hline
GoogLeNet          & {10-8-10-9-8-10-9-8-9-10-7}                       & {11}                   \\ 
\hline
VGGS               & {7-8-9-7-9}                                       & {12}                   \\ 
\hline
VGGM               & {7-7-7-8-7}                                       & {12}                  \\ 
\hline
AlexNet-Sparse~\cite{cvpr_2017_yang_energy }          & {8-9-9-9-8}    & {7}         \\ 
\hline
ResNet50-Sparse~\cite{SkimCaffePaper}               & {10-8-6-6-5-7-6-6-7-7-6-7-6-7-6-8-7-6-8-6-5-8
-8-6-8-7-7-6-9-7-5-8-7-6-8-7-6-8-7-6-8-8-7-8-7-9-6-10-7-6-10-8-7}  & {13}         \\ 
\hline

\end{tabular}
\label{tab:CONV_precisions}
\end{table*}

Figure~\ref{fig:performance} shows the performance of $\OURS$ configurations relative to $\BASE_{2K}$ for convolutional layers with the 100\% relative TOP-1 accuracy precision profiles of Table~\ref{tab:CONV_precisions}. 

\OURL targets both dense and sparse networks and improves the performance by processing only the essential terms; however, the sparse networks would benefit more as they posses more ineffectual terms. On average, $\OURS_{128}$ outperforms $\BASE_{2K}$ by more than $2\times$ while for AlexNet-Sparse $\OURS_{128}$ achieves a speedup of $3\times$ over the baseline. Figure~\ref{fig:performance} shows how average performance on convolutional layers over all networks scales for different configurations with different number of weight wires. $\OURS_{256}$, $\OURS_{512}$, and $\OURS_{1K}$ achieve speedups of $4.0\times$, $8.1\times$, and $15.4\times$ over $\BASE_{2K}$, respectively.

\subsection{Energy Efficiency}
\label{sec:e:e}

\begin{table}
    \centering
    \footnotesize
    \caption{
\OURL energy efficiency relative to $\BASE_{2K}$. 
}
	\begin{tabular}{|l|l|l|l|l|} 
\hline
 & $\OURS_{128}$  & $\OURS_{256}$ & $\OURS_{512}$ & $\OURS_{1K}$\\ \hline                   
AlexNet & 2.03 & 2.44 & 2.92 & 1.88\\ \hline
GoogLeNet & 1.84 & 2.32 & 2.76 & 1.75\\ \hline
VGG\_S & 2.43 & 3.04 & 3.63 & 2.31 \\ \hline
VGG\_M & 2.18 & 2.73 & 3.26 & 2.02\\ \hline
AlexNet-Sparse & 2.81 & 2.91 & 3.49 & 2.19\\ \hline 
ResNet-Sparse & 1.69 & 2.02 & 2.41 & 1.61\\ \hline \hline
Geomean & 2.13 & 2.55 & 3.05 & 1.95 \\
\hline
\end{tabular}
\label{tab:energy}
\end{table}

Table~\ref{tab:energy} summarizes the energy efficiency of various $\OURS$ configurations over $\BASE_{2K}$. On average over all networks $\OURS_{128}$, $\OURS_{256}$, $\OURS_{512}$, and $\OURS_{1K}$ are $2.13\times$, $2.55\times$, $3.05\times$, and $1.95\times$ more energy efficient than $\BASE_{2K}$.

\subsection{Area}
\label{sec:e:a}

Post layout measurements were used to measure the area of \BASE and \OURS. The $\OURS_{128}$, $\OURS_{256}$, and $\OURS_{512}$ configurations require $0.75\times$, $0.82\times$, and $0.96\times$ less area than $\BASE_{2K}$, respectively while outperforming $\BASE_{2K}$ by $2.3\times$, $4.0\times$, and $8.1\times$. The area overhead for $\OURS_{1K}$ is $1.36\times$ while its execution time improvement over the baseline is $15.4\times$. Thus \OURS exhibits better performance vs. area scaling than \BASE. 


\subsection{Scalability}
Thus far we considered designs with up to 1K wire weight memory connections. For one of the most recent network studied here, GoogleNet we also experimented with 2K and 4K wire configurations. Their relative performance improvements were $20.4\times$ and $27.0\times$. Similarly to other accelerators performance improves sublinearly. This is primarily due to inter-filter imbalance which is aggravated as in these experiments we considered only increasing the number of filters when scaling up. Alternate designs may consider increasing the number of simultaneously processed activations instead. In such configurations, minimal buffering across activation columns as in Pragmatic~\cite{pragmatic} can also combat cross-activation imbalance which we expect will worsen as we increase the number of concurrently processed activations.




\section{Conclusion}
\label{sec:theend}

We have shown that compared to conventional bit-parallel processing,  aiming to process only the non-zero bits (or terms in a booth-encoded format) of the activations and weights has the potential to reduce work and thus improve performance by two orders of magnitude. We presented the first practical design, \OURL that takes advantage of this approach leading to best-of-class performance improvements. \OURL is naturally compatible with the compression approach of Delmas et al.,~\cite{DPRed} and thus as per their study we expect to perform well with practical off-chip memory configurations and interfaces. 



\bibliographystyle{ieeetr}
\interlinepenalty=10000
\bibliography{ref2}

\end{document}